\documentclass[
twocolumn,
]{ceurart}

\usepackage{listings}
\begin{document}

\copyrightyear{2024}
\copyrightclause{Copyright for this paper by its authors.
  Use permitted under Creative Commons License Attribution 4.0
  International (CC BY 4.0).}

\conference{CLiC-it 2024: Tenth Italian Conference on Computational Linguistics, Dec 04 — 06, 2024, Pisa, Italy}

\title{Harnessing LLMs for Educational Content-Driven Italian Crossword Generation}

\author[1]{Kamyar Zeinalipour}[%
orcid=0009-0006-3014-2511,
email=kamyar.zeinalipour2@uisi.it,
url=https://kamyarzeinalipour.github.io.,
]
\cormark[1]
\fnmark[1]
\address[1]{University of Siena, DIISM, Via Roma 56, 53100 Siena, Italy}

\author[2]{Achille Fusco}[%
orcid=0000-0002-5389-8884,
email=achille.fusco@iusspavia.it,
]
\fnmark[1]
\address[2]{IUSS Pavia, Piazza della Vittoria 15,
27100 Pavia (PV)}

\author[1]{Asya Zanollo}[%
email=a.zanollo@student.unisi.it
]
\fnmark[1]

\author[1]{Marco Maggini}[%
orcid=00000-0002-6428-1265,
email=marco.maggini@unisi.it
]

\author[1]{Marco Gori}[%
orcid=0000-0001-6337-5430,
email= marco.gori@unisi.it
]
\cortext[1]{Corresponding author.}
\fntext[1]{These authors contributed equally.}

\begin{abstract}
In this work, we unveil a novel tool for generating Italian crossword puzzles from text, utilizing advanced language models such as \texttt{GPT-4o}, \texttt{Mistral-7B-Instruct-v0.3}, and \texttt{Llama3-8b-Instruct}. Crafted specifically for educational applications, this cutting-edge generator makes use of the comprehensive \textit{Italian-Clue-Instruct} dataset, which comprises over 30,000 entries including diverse text, solutions, and types of clues. This carefully assembled dataset is designed to facilitate the creation of contextually relevant clues in various styles associated with specific texts and keywords.
The study delves into four distinctive styles of crossword clues: those without format constraints, those formed as definite determiner phrases, copular sentences, and bare noun phrases. Each style introduces unique linguistic structures to diversify clue presentation.
Given the lack of sophisticated educational tools tailored to the Italian language, this project seeks to enhance learning experiences and cognitive development through an engaging, interactive platform. By meshing state-of-the-art AI with contemporary educational strategies, our tool can dynamically generate crossword puzzles from Italian educational materials, thereby providing an enjoyable and interactive learning environment. This technological advancement not only redefines educational paradigms but also sets a new benchmark for interactive and cognitive language learning solutions.
\end{abstract}


\maketitle

\section{Introduction}

While traditionally valued for their challenge and entertainment, crossword puzzles are increasingly recognized for their educational benefits. They provide an interactive learning environment that enhances the retention of both technical terms and general language skills, hence facilitating learning across various disciplines, improving language acquisition, and supporting cognitive development, through critical thinking and memory retention  \cite{orawiwatnakul2013crossword,bella2023improving,dzulfikri2016application,nickerson1977crossword,yuriev2016crossword,sandiuc2020use,kaynak2023effect,dzulfikri2016application,mueller2018testing,zirawaga2017gaming,bella2023improving,zamani2021use,dol2017gpbl}.\\
The integration of Natural Language Processing (NLP) and Large Language Models (LLMs) has further enhanced their effectiveness by providing sophisticated, contextually relevant clues for educational crosswords.\\
This paper presents a novel tool that uses LLMs to generate tailored Italian educational crossword puzzles from texts, offering various clue types. By integrating user-provided texts or keywords and applying fine-tuning techniques, the tool produces high-quality clues and answers, offering educators a resource to develop more interactive and effective instructional methods.\\
Furthermore, a new dataset called \footnote{\url{https://huggingface.co/datasets/Kamyar-zeinalipour/ita_cw_text}}
  has been compiled and will be released to the scientific community. \\
The layout of this paper is organized in the following manner: Section \ref{sec:relatedworks} surveys the relevant literature in detail. Section \ref{sec:dataset} explains the methods used for dataset collection and curation. In Section \ref{sec:Methodology}, we describe the computational techniques employed in our study. Section \ref{sec:Experiments} reports the results derived from our experimental analysis. Finally, Section \ref{sec:conclusions} closes with conclusive insights and the broader implications of our research findings.

\section{Related Works}
\label{sec:relatedworks}
Among the pioneering efforts in the field of crossword puzzle generation, Ranaivo et al. have formulated a distinctive strategy that merges text analytics with graph theory, allowing for the extraction and refinement of topic-specific clues through NLP~\cite{ranaivo2013automatic}. Another notable contribution comes from Rigutini et al., who laid the groundwork by utilizing advanced NLP to automatically generate crossword puzzles from online sources, representing a seminal step in the field~\cite{rigutini2008fully, rigutini2012automatic}.\\
In parallel, Esteche and his team have focused on Spanish-speaking audiences by creating puzzles with the aid of electronic dictionaries and news articles to formulate clues~\cite{esteche2017automatic}. \\
On a different front, Arora et al. developed SEEKH, a system that integrates statistical and linguistic analyses to generate crossword puzzles in multiple Indian languages. Their approach emphasizes the identification of keywords to structure the puzzles~\cite{arora2019automatic}.\\
Recent progress in crossword puzzle generation has been notably advanced by the work of Zeinalipour et al. \cite{zeinalipour2023building,zeinalipour2023italian,zeinalipour2023arabicros,zeinalipour2024turkish}, who demonstrated the use of large-scale language models to develop puzzles in languages with limited support, such as English, Italian and Arabic. Their research highlights the vast potential of computational linguistics in crafting puzzles that are both engaging and linguistically rich. Initially, they employed few-shot and zero-shot learning techniques to generate new crossword clues from text ~\cite{zeinalipour2023italian, zeinalipour2023building}.\\
Furthermore, Zugarini et al. \cite{zugarini2024clue} introduced a method for generating educational crossword clues from the provided text in English.\\
In their Italian crossword puzzle generation study~\cite{zeinalipour2023italian}, Zeinalipour et al. initially used few-shot learning with large language models as-is. However, our current project goes a step further by introducing a specially designed dataset for this task in Italian. Additionally, we have developed open-source models that have been fine-tuned to significantly enhance performance for this specific application.\\
The current research initiates a novel approach by utilizing state-of-the-art language modeling to develop Italian crossword puzzles from given texts. By doing so, it enriches the toolkit for language education, thereby pushing forward the development of Italian crossword puzzles.

\section{Methodology}\label{sec:Methodology}
We have developed an automated system that generates educational Italian crossword puzzles using LLMs, with the \textit{Italian-Clue-Instruct} dataset at its core. Our approach leverages the adaptability of LLMs, like GPT-4o, to create puzzles from text, with human validation for accuracy. Additionally, we fine-tuned models such as \texttt{Llama3-8b-Instruct} and \texttt{Mistral-7B-Instruct-v0.3} to improve clue accuracy and relevance.

\begin{figure*}[ht!]
    \centering
       \includegraphics[width=\textwidth]{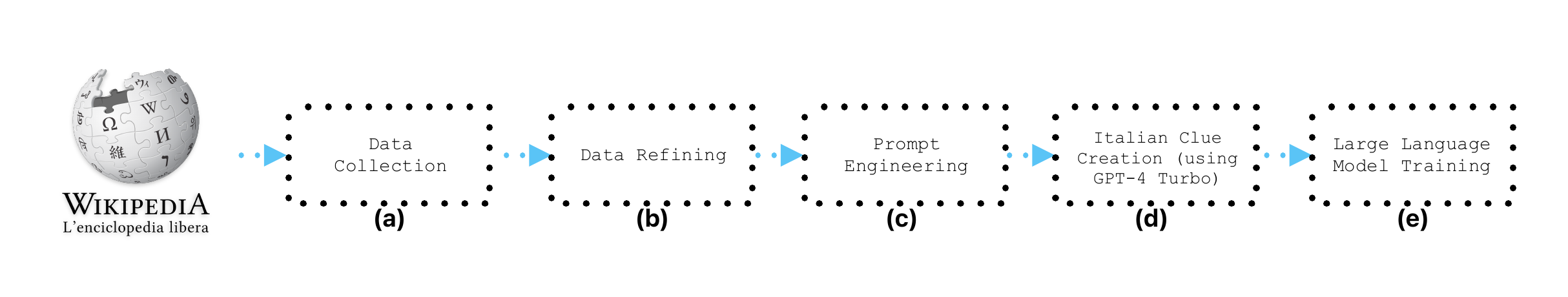}
    \caption{        The methodology followed in this study comprises the following stages: 
        (a) Gathering an extensive dataset from the Italian Wikipedia.
        (b) Refining and filtering the data by eliminating entries that are either too brief or excessively detailed, thereby optimizing its quality.
        (c) Developing specialized prompts intended to create educational Italian crossword clues derived from the curated dataset.
        (d) Utilizing \texttt{GPT-4o} to generate Italian crossword clues based on the processed data and crafted prompts.
        (e) Fine-tuning Large Language Models (LLMs) to enhance their performance in producing contextual and tailored Italian crossword clues. 
        These systematic steps ensure the effective leveraging of advanced natural language processing technologies to create high-quality educational content in the form of Italian crossword clues.}
    \label{fig:fig1}
\end{figure*}

A more detailed description of our methodology, illustrated in Figure \ref{fig:fig1}, is provided in the following. 

\paragraph{\textit{Italian-Clue-Instruct}} 
\label{sec:dataset}

\paragraph{Data Collection Methodology} \label{sec:data_acquisition}
Initiating the data collection process, we began by extracting the introductory portions of Italian Wikipedia articles. We use Wikipedia API and Beautiful Soup to automatically extract the pages. The prominent focus was placed on the bolded keywords that highlight the primary topic and other significant terms within each article. Beyond keyword identification, we also gathered a variety of essential metadata. This included metrics such as view counts, relevance assessments, brief narrative summaries, central headlines, related terms, categorization, and URLs.\footnote{\href{https://en.wikipedia.org/wiki/Wikipedia:Lists_of_popular_pages_by_WikiProject}{Wikipedia: Lists of popular pages by WikiProject}}
The uniform structure of the Italian Wikipedia significantly aids this process.  By tapping into the introductory sections, which are particularly information-rich, we could systematically extract and outline the key concepts needed. This approach ensures a comprehensive data repository, capturing critical elements and insights from a diverse array of articles.

\paragraph{Data Enhancement} \label{sec:data_filtering}
To ensure the reliability and effectiveness of our data, we performed some filtering based on different criteria. The first filter was designed to prioritize the most important pages and those with the highest number of views.
Firstly, articles were selected based on their popularity and relevance. To ensure a balanced and manageable dataset, we also discarded articles that were either too lengthy or too brief, specifically those with fewer than 50 words. Additionally, we removed keyword associations longer than two words to maintain the clarity and relevance of the crossword clues. Finally, we imposed restrictions on keywords to ensure they were between 3 and 20 characters in length and free of special characters or numerals. Multi-words expressions were also included as good keywords as they are quite common in crossword puzzles.

\paragraph{Formulation of Various Prompts} \label{sec:datprompt}
Crafting specialized prompts was pivotal for producing Italian crossword clues from a given text using GPT-4o. The prompts were created to generate clues that were both informative and engaging, by incorporating crucial details and background context from the articles.
Additionally, apart we aimed to elicit three specific types of clue varying in their syntactic structures: 
\begin{itemize}
    \item \texttt{definite determiner phrases}: nominal clues headed by a definite article and usually modified by adjectives, prepositional phrases (PPs) or relative clauses (RCs), like <\textit{La repubblica asiatica con capitale Tashkent}, \textit{Uzbekistan}> (`The Asian republic with Tashkent as capital', `Uzbekistan'). Such clues are examples of definite descriptions which have been traditionally analyzed as carrying a uniqueness presupposition (\cite{chierchia1998reference}) when singular and a maximality presupposition \cite{link1983logical} when plural. In the context of crosswords, clues of this kind refer to their solution as the single entity or the maximal plural entity satisfying the description.
    \item \texttt{bare noun phrases} \cite{longobardi1994reference}: the clue consists of a simple noun phrase (NP) with no determiner and typically modified by adjectives, PPs or RCs, for example <\textit{Grande centro commerciale di lusso con sede a Londra, Harrods}> (`Luxury shopping mall based in London', `Harrods'). In Italian, NPs are taken to denote a predicate that can be true of one or more individuals \cite{chierchia1998reference, zamparelli2014layers}.\footnote{Bare NPs are known to denote also natural kinds \cite{chierchia1998reference}. However, given that NP clues occur in isolation, it is rather difficult to distinguish among the two senses, therefore we assume the more general reading of NPs as predicates. We leave this discussion to future analyses.} Given the absence of the definite determiner, bare NP clues do not specify whether the referent of the solution uniquely satisfies the description \cite{chierchia1998reference}, thus more than one solution could in principle be possible. 
    \item \texttt{copular sentences} \cite{moro2006copular}: copular clues are clausal definitions structured as <\textit{copula predicate}> with an elliptical subject as in <\textit{è una salsa piccante tipica della Tunisia, Harissa}> (`(It) is a spicy sauce typical of Tunisia', `Harissa'). Copulas, like Italian \textit{essere} ('to be') connect a subject with a non-verbal predicate, such as an adjectival phrase (AP), a PP or another nominal phrase (NP/DP). In crossword puzzles, the solution targets the precopular position of such sentences, i.e. the elliptical subject. \footnote{Copular sentences are known to be differentiated between canonical and inverse structures \cite{moro2006copular}. Usually in crossword clues canonical structure are found more frequently, but inverse copular clues are not excluded. We leave the question open for further, purely linguistic research.}
\end{itemize}
To accomplish this, we created three distinct prompts for each clue structure, and one prompt that does not specify the structure. 
This step allows us to test the syntactic sensitivity of the models employed and, more importantly it gives us the possibility of manipulating the structure to create variation not just with respect to the subject matter but also in the clue syntactic complexity. Moreover, generating clues with specific structures represents an interesting resource for the educational characterization of puzzles. Indeed, it is well-known from psycholinguistic research that different structures can elicitate different reactions in the processing which can be correlated with factors like age, linguistic disorders etc. and this can be exploited when creating puzzles specific for any solver's needs. 

As for the prompt engeneering, the structure has been explicitated in one dedicated step of the prompt chain. For what regards the copular structure, which is widespread and widely used with different formulation, we include an example in the prompt (as shown in \ref{fig:prompt4}) to ensure that the required structure is given in output. 
It has been observed during the prompt trials that the validity of precise structures for clues strongly depends on the type of text given in input. The prompts used for clue generation in this study are presented in Figures  \ref{fig:prompt}, \ref{fig:prompt2}, \ref{fig:prompt3} and \ref{fig:prompt4}, located in the Appendix.\\

\paragraph{Generation of Educational Italian Clues.} \label{sec:datagen}
Guided by the \textsc{self-instruct} framework~\cite{wang2022self}, we devised a method to automate the generation of educational crossword clues in Italian, harnessing the power of LLMs. Central to our approach is the sophisticated \texttt{GPT-4o}\footnote{https://openai.com/index/hello-gpt-4o/}, an enhanced version of LLMs, renowned for its efficiency. A key differentiator of our strategy is the integration of contextual information with the clues produced. To achieve this, we carefully curated the content and keywords from the Wikipedia text extracted in previous sections. We used four distinct types of prompts, each designed to generate different categories of clues: bare noun phrases, definite determiner phrases, and copular sentences. These prompts were crafted to create diverse types of clues, ensuring alignment with our specific objectives for educational content in Italian.

\paragraph{Overview of the \textit{Italian-Clue-Instruct} Dataset}

Our research began with downloading 88,403 articles from the Italian Wikipedia, which we filtered down to 11,413 relevant entries. From this refined set, we selected 5,000 articles for clue generation, spanning 29 thematic categories.  
To enhance our dataset, we leveraged the capabilities of \texttt{GPT-4o}, generating a minimum of three diverse clues per Wikipedia article, depending on the text length. This effort resulted in a compilation of 15,000 unique clues.

\begin{figure*}
    \centering
       \includegraphics[width=\textwidth]{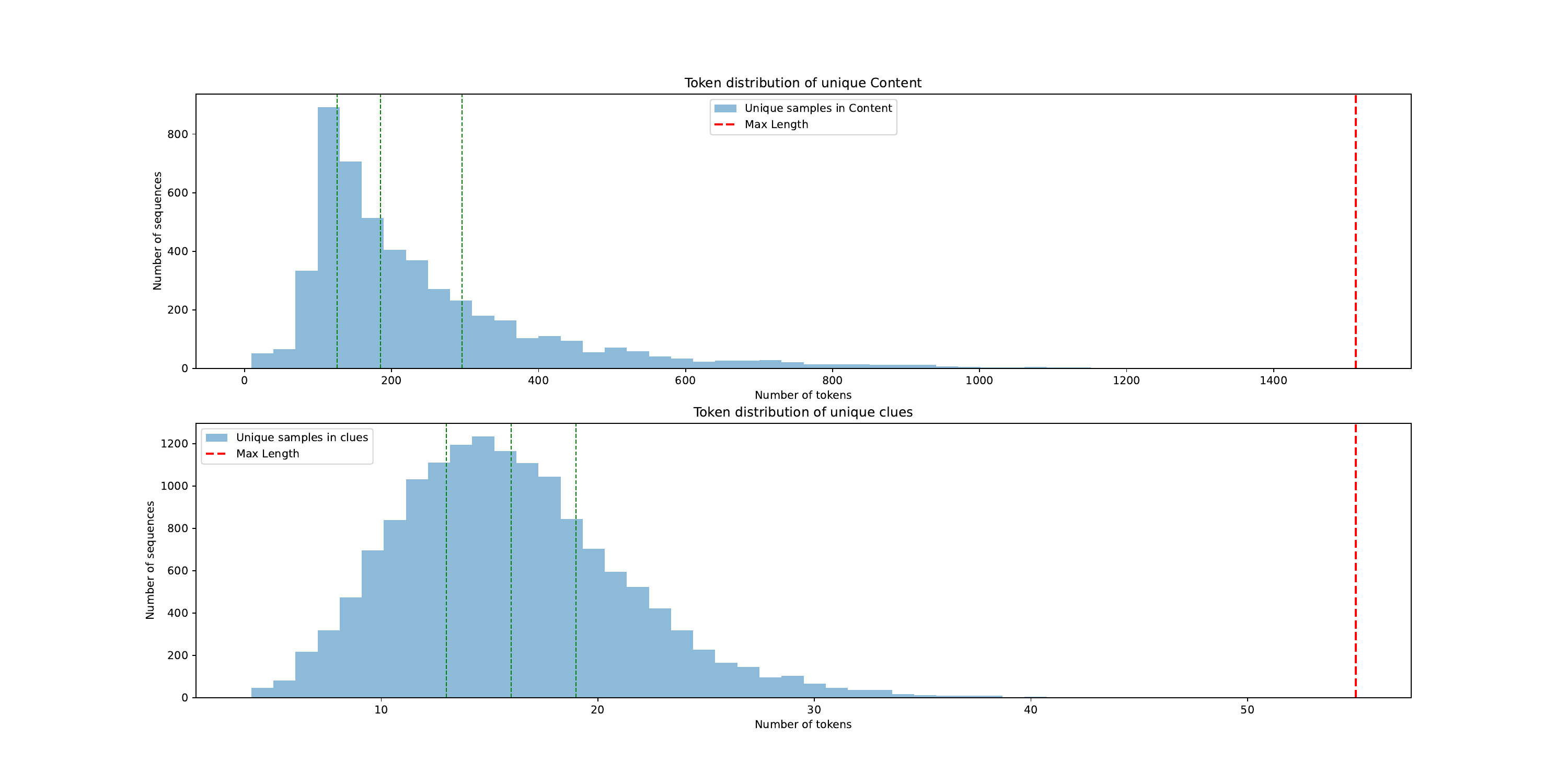}
    \caption{Token Distributions for Context and Clues of \textit{Italian-Clue-Instruct}}
    \label{fig:dataset_distrubtions}
\end{figure*}

\begin{figure*}
    \centering
       \includegraphics[width=\textwidth]{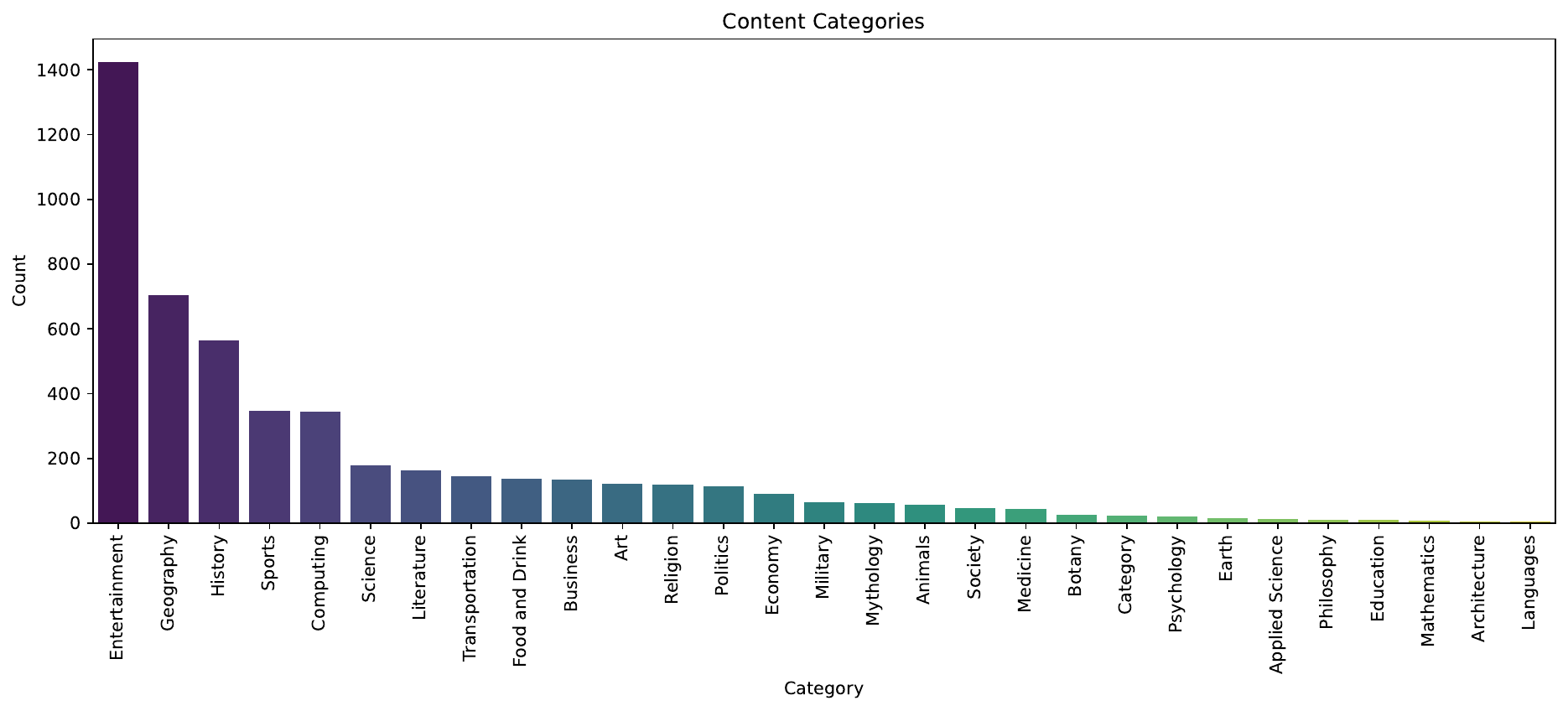}
    \caption{Bar Plot Showing the Frequency of Different Categories within the Dataset.}
    \label{fig:topics-distr}
\end{figure*}

The dataset's in-depth analysis demonstrates a variability in context length, ranging from 10 to 1512 tokens, with most texts falling between 100 and 600 tokens. Figure \ref{fig:dataset_distrubtions} showcases the token distribution for contexts and clues, which have been processed using the Llama3 tokenizer. Typically, the clue-generation process results in clues ranging from 4 to 55 tokens in length.\\
Figure \ref{fig:topics-distr} illustrates the spread of data across different categories. The dataset is notably dominated by the categories of "Entertainment", "Geography", and "History". In contrast, categories such as "Mathematics", "Architecture", and "Languages" are underrepresented.

\paragraph{Evaluating quality of the \textit{Italian-Clue-Instruct} Dataset}

Producing accurate and engaging Italian educational crossword clues is inhibited by the absence of a reference corpus, making it difficult to draw comparisons using standard measures, such as ROUGE scores. Our evaluation strategy adapts uniquely to the task requirements. Specifically, effective clues should represent contextually accurate paraphrases of text information. To accommodate this, we adopted an extractive method, using the ROUGE-L score to gauge the adequacy of clues in reflecting the input context that we extracted from Wikipedia. By comparing input sentences to the generated clues, the evaluation aimed to attain high scores to ensure strict adherence to the original text, minimizing irrelevant content and avoiding clues that merely replicate the input or improperly introduce the target keyword. Results indicated a substantial connection between the context and the clues, with an average ROUGE-1, ROUGE-2, and ROUGE-L score of 0.159, 0.114, and 0.146 respectively.

Considering that the ROUGE score merely compares the similarity between the n-grams of the generated clues and the reference text from Wikipedia, it is not a reliable metric and does not provide any assessment of the semantic quality of the generated clues. However, it provides a general picture of the generated clues.\\
In addition, the integrity of the generated clues was further examined through human evaluations. A randomly chosen subset of clues was assessed, generated from a sample of 100 articles, with a maximum of three clues per article. To avoid repetitions, duplicate clues were removed. The evaluation employed a five-level criteria system, analogous to the methodology utilized by \cite{wang2022self}. For the present evaluation, the following parameters were used:
\begin{itemize}
    \item \texttt{RATING-A}: The clue is coherent and valid, aligning correctly with the given context, answer, and specified structure.
    \item \texttt{RATING-B}: This clue, while generally acceptable, exhibits slight discrepancies mainly due to sub-optimal phrasing or structure.
    \item \texttt{RATING-C}: The clue relates directly to the answer but retains a vague connection to the context or provide information which, even if correct, is not properly conveyed.
    \item \texttt{RATING-D}: The clue is strictly referring to the context and fails to comprehensively identify the answer.
    \item \texttt{RATING-E}: The clue is deemed unacceptable because it is ungrammatical, it directly contains the answer or a variation of it, or doesn't identify the referent of the answer.
\end{itemize}

The evaluation was made by a native Italian speaker, master student of linguistics, and PhD student in linguistics, who followed the criteria described above. Please refer to Table \ref{tab:examples} for examples of clues and their respective ratings.  

\begin{figure}
    \centering 
   \includegraphics[width=\columnwidth]{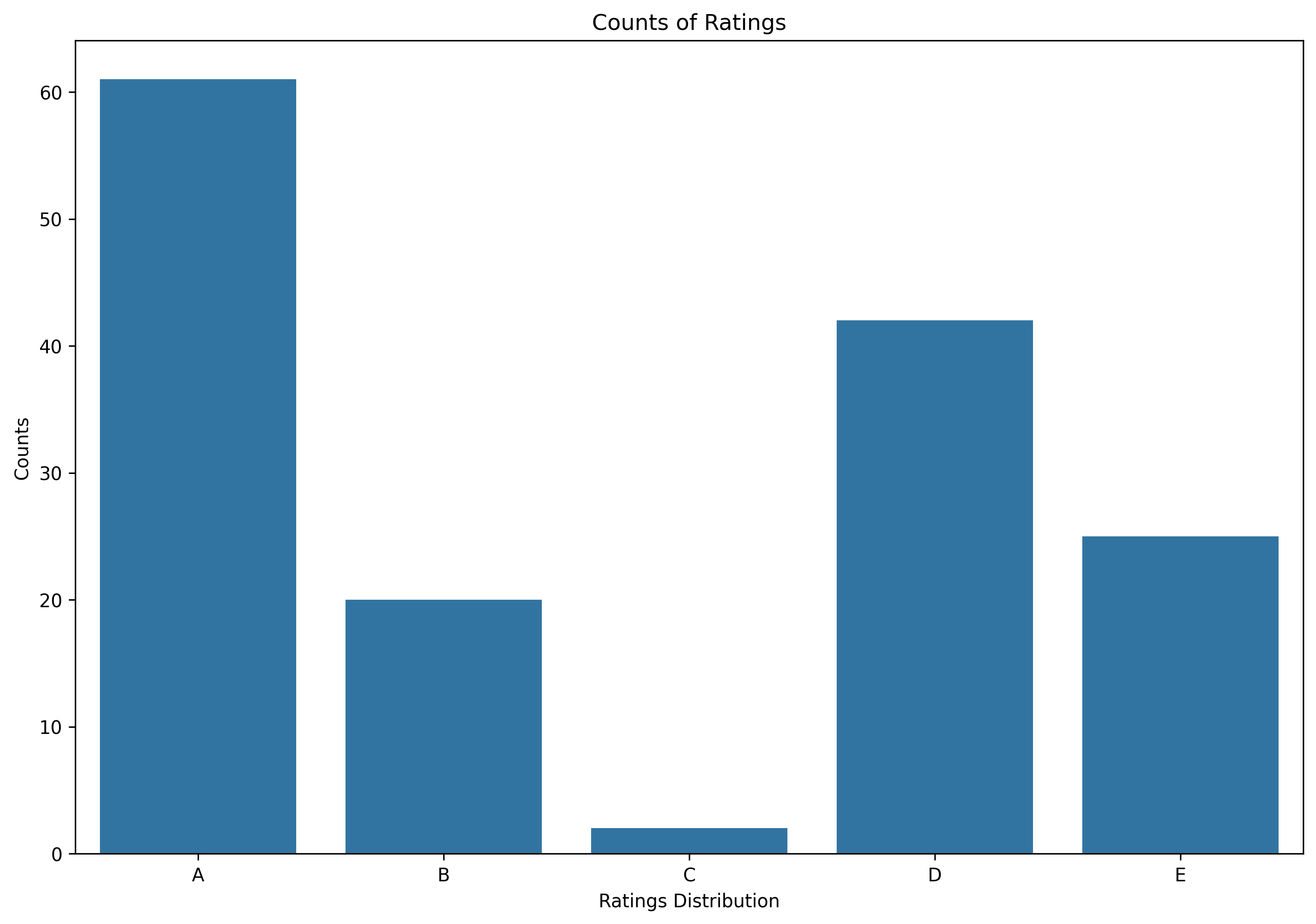}
    \caption{Bar Plot Showing the Frequency of GPT-4o Ratings}
    \label{fig:count_of_ratings}
\end{figure}

The distribution of the evaluation outcomes is depicted in Figure \ref{fig:count_of_ratings}, these illustrate that the majority of the generated clues were of high quality rated as 'A' and only a small fraction rated as 'C', 'D', or 'E'.\\

By utilizing both quantitative metrics and qualitative assessments, the study aimed to validate the educational utility and contextual accuracy of the clues created for Italian educational crosswords.

\paragraph{Enhancing LLMs for Italian text-based Educational Crossword Puzzle Generation}

To develop crossword puzzle clues from Italian texts using advanced LLM functionalities, we employed three models: \texttt{GPT-4o} (for data generation), \texttt{Mistral-7B-Instruct-v0.3}, and \texttt{Llama3-8b-Instruct} known for their strong text generation and Italian language support. \cite{brown2020language,touvron2023llama}.\\ 
We began the process by fine-tuning the models with the \textit{Italian-Clue-Instruct} dataset, which was rich in relevant material. This calibration was vital to enhance the models' proficiency in generating Italian clues while accurately reflecting the Italian language's intricate grammar and vocabulary within educational contexts.\\
To further refine the models, we optimized the parameters during the fine-tuning phase. This effort aimed to reduce errors specific to our task and better align the output of the models with Italian educational materials.\\
Ultimately, the specialized tuning of these LLMs with a dedicated dataset was intended to foster their ability to generate high-quality crossword clues from Italian texts. The goal was to ensure that the resulting clues were not only linguistically sound but also relevant within an educational framework.

\section{Experimental Results}\label{sec:Experiments}
This section offers a detailed overview of the experiments conducted in the study. It begins with the training setup for the \textit{Italian-Clue-Instruct} LLMs, including key parameters and computational resources. The performance of the models is then evaluated using automated metrics, such as the ROUGE score, to compare configurations and identify areas for improvement. This is followed by an in-depth analysis of human evaluations, focusing on relevance, coherence, and content quality to provide insights beyond automated metrics. Additionally, an example of a generated crossword puzzle is presented to demonstrate practical usability. The goal is to highlight the robustness and versatility of the proposed approach. 
\paragraph{Training Setup}
The models \texttt{Mistral-7B-Instruct-v0.3} and \texttt{Llama3-8b-Instruct} were fine-tuned using LORA \cite{hu2021lora}, with parameters set to $r=16$ and $\alpha=32$, across three training epochs, maintaining a total batch size of 64. The full experimental setup was performed on a server equipped with four NVIDIA A6000 GPUs, utilizing DeepSpeed \cite{rasley2020deepspeed} and FlashAttention 2 \cite{dao2023flashattention}. For the initial learning rate was configured at $3\times 10^{-4}$. During inference, model distribution sampling was applied to generate clues for both \texttt{Mistral-7B-Instruct-v0.3} and \texttt{Llama3-8b-Instruct}, with a temperature parameter set to 0.1. Additionally, the parameters for top-$p$ and top-$k$ sampling were set to 0.95 and 50, respectively. Among the three epoch checkpoints, the one with the minimum loss was selected, which, in our case, turned out to be the second checkpoint.

\paragraph{Evaluation Results with the Automatic Metrics}

We evaluated the resemblance between various sets of clues produced by different models (details shown in Table \ref{tab:rouge_scores}) and those generated by the \texttt{GPT-4o} model on a test set of 200 educational contexts. This evaluation was done using ROUGE scores. Our results indicate that the fine-tuned \texttt{Mistral-7B-Instruct-v0.3} and \texttt{Llama3-8b-Instruct} models exhibit a closer similarity to \texttt{GPT-4o}. On the other hand, the base \texttt{Llama3-8b-Instruct} model shows significantly lower similarity with minimal overlap. These outcomes highlight the efficacy of fine-tuning, demonstrating that using the \textit{Italian-Clue-Instruct} dataset enhances the capability of \texttt{Mistral-7B-Instruct-v0.3} and \texttt{Llama3-8b-Instruct} models in generating clues from Italian educational texts.

\paragraph{Evaluation Results with the human evaluator }

Using a dataset of 100 Italian contexts, each containing 3 clues, a human evaluation was conducted on both the generated and base models. The results of this evaluation are depicted in Figure \ref{fig:ratings-dist}. The evaluation employed the 5-level rating system described in Section \ref{sec:Methodology}.

The table provided offers a comparative evaluation of the performance of language models in generating Italian clues from a given text. Specifically, the models \texttt{Mistral-7B-Instruct-v0.3} and \texttt{Llama3-8b-Instruct} are evaluated based on both their base and fine-tuned configurations. Upon fine-tuning, \texttt{Mistral-7B-Instruct-v0.3} displays a significant improvement, emerging as the top performer in category "A", and surpassing \texttt{Llama3-8b-Instruct} in terms of performance enhancement. These findings underscore the impact of fine-tuning on enhancing model capabilities, particularly highlighted by the performances of \texttt{Mistral-7B-Instruct-v0.3} and \texttt{Llama3-8b-Instruct}, which feature 7 and 8 billion parameters, respectively. Furthermore, fine-tuning with the introduced dataset significantly increased the models' ability to generate Italian clues from the given text, illustrating the quality and effectiveness of the \textit{Italian-Clue-Instruct} dataset.\\
The methodology for generating Italian crossword clues from educational texts was explored, enabling customized clues. This would allow educators to select suitable clues matching their teaching needs. The selected clues could in turn be used to automatically generate a crossword schema as discussed \citet{zeinalipour2023building}. Figure \ref{fig:crossword} in Appendix shows an example puzzle, demonstrating the system's application.

\section{Conclusion}\label{sec:conclusions}

A novel system for generating crossword clues from Italian text is introduced, leveraging the newly developed \textit{Italian-Clue-Instruct} dataset. This dataset, which includes text, keywords, categories, and related crossword clues in Italian, is pioneering in this field. By fine-tuning two large language models (LLMs), \texttt{Mistral-7B-Instruct-v0.3} and \texttt{Llama3-8b-Instruct}, using this dataset, we have achieved significant improvements in the models' ability to generate crossword clues from given text. The results highlight a substantial enhancement in model performance after fine-tuning. Both the \textit{Italian-Clue-Instruct} dataset and the fine-tuned models are now publicly available, providing valuable tools for students and teachers to create educational crossword puzzles from Italian text. Future research will aim to develop models capable of generating various types of crossword clues, including fill-in-the-blank clues.

\begin{acknowledgments}

The funding for this paper was provided by the TAILOR project and the HumanE-AI-Net projects, both supported by the EU Horizon 2020 research and innovation program under GA No 952215 and No 952026, respectively.

\begin{table*}[ht]
    \centering
\begin{tabular}{c c c c c} 
 \hline
 \textbf{Model} & \textbf{Model name} & \textbf{ROUGE-1} & \textbf{ROUGE-2} & \textbf{ROUGE-L} \\
 \hline
  	Base LLMs & Mistral-7B & 0.342 & 0.176 &  0.261 \\

        &Llama3-8b& 0.258 & 0.112& 0.198 \\
  \hline
 
	Fine-tuned LLMs  & Mistral-7B &\textbf{0.611} &\textbf{0.458}  &\textbf{0.556} \\ 

          & Llama3-8b &  0.552 &  0.403& 0.501 \\
    \hline
\end{tabular}
    \caption{Mean ROUGE Scores for Various Comparisons with GPT-4o generated clues}
     \label{tab:rouge_scores}
\end{table*}
\begin{figure*}[hbt!]
    \centering 
   \includegraphics[width=\textwidth]{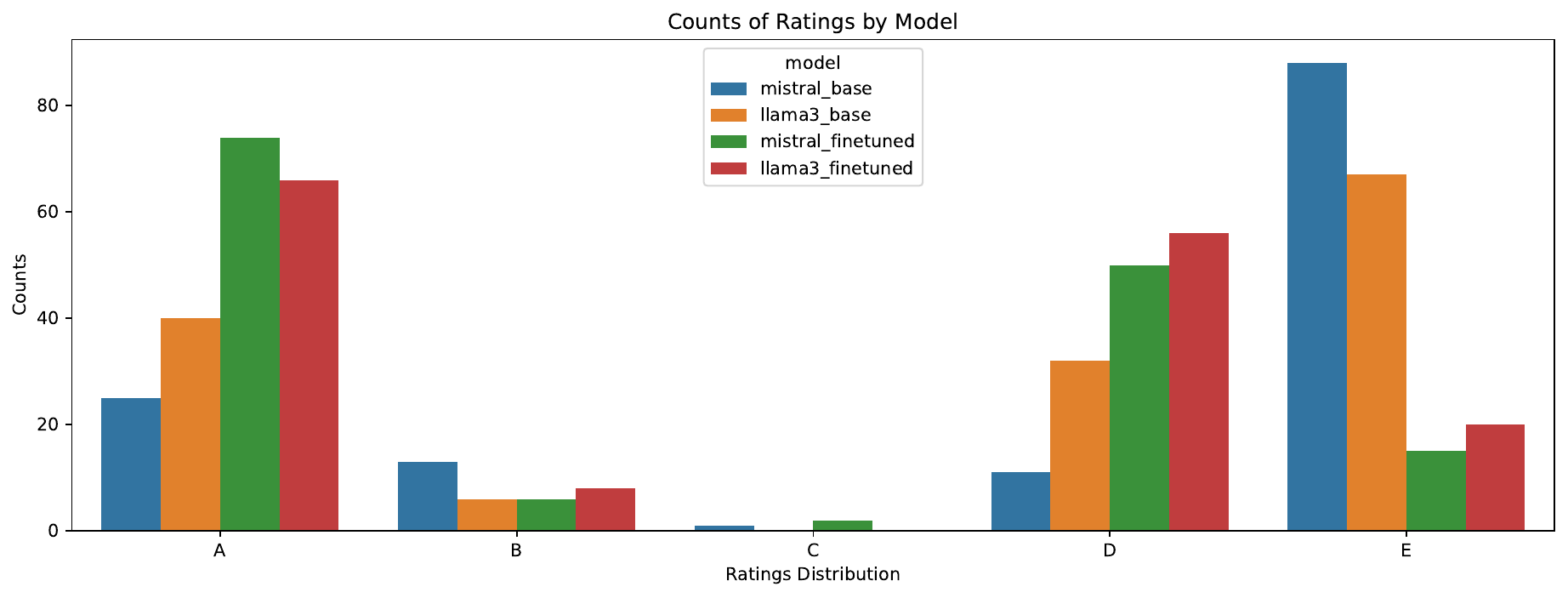}
    \caption{Bar Plot Showing the Frequency of the ratings after the evaluation.}
    \label{fig:ratings-dist}
\end{figure*}

\end{acknowledgments}

\bibliography{sample-ceur}

\appendix

\section{Appendix}
\begin{figure*}[ht!]
    \centering
    \includegraphics[width=\textwidth]{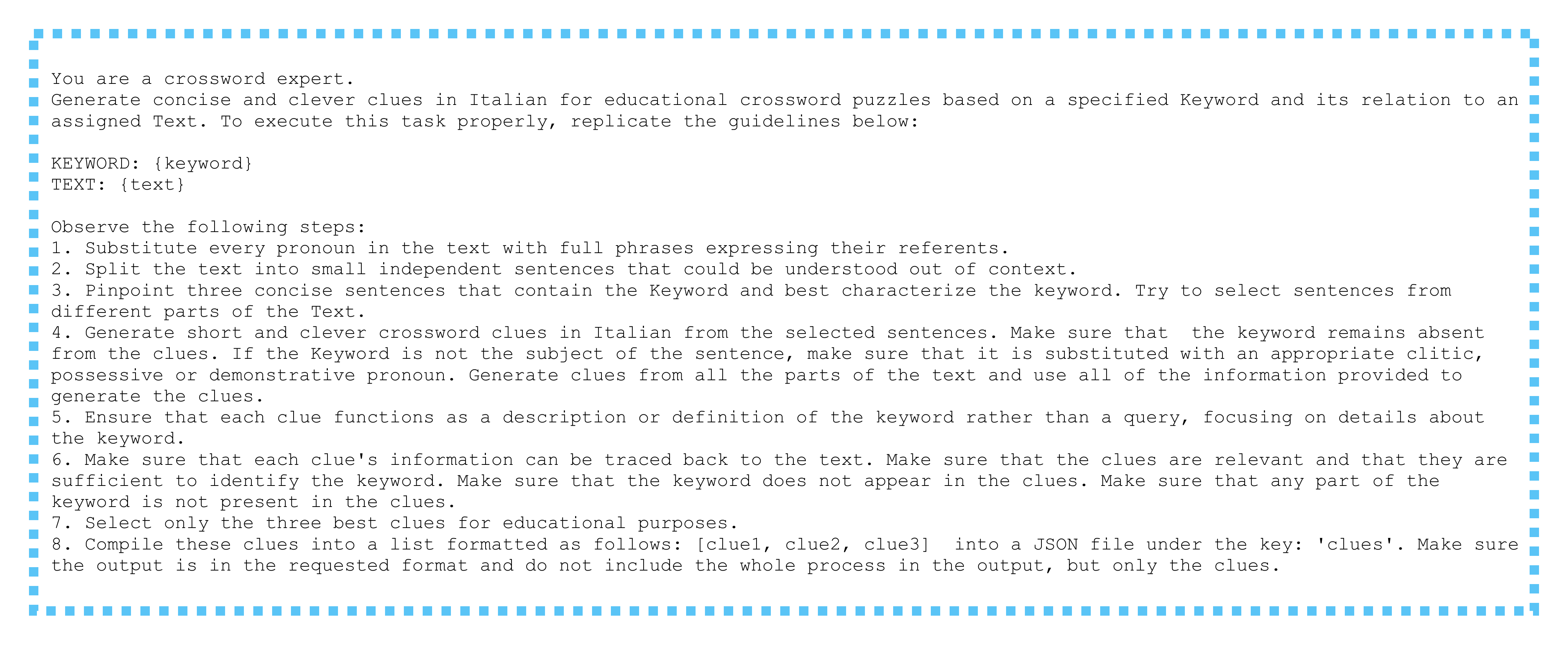}
    \caption{Illustration of the prompt used for unrestricted format clues in the research.}
    \label{fig:prompt}
\end{figure*}

\begin{figure*}[ht!]
    \centering
    \includegraphics[width=\textwidth]{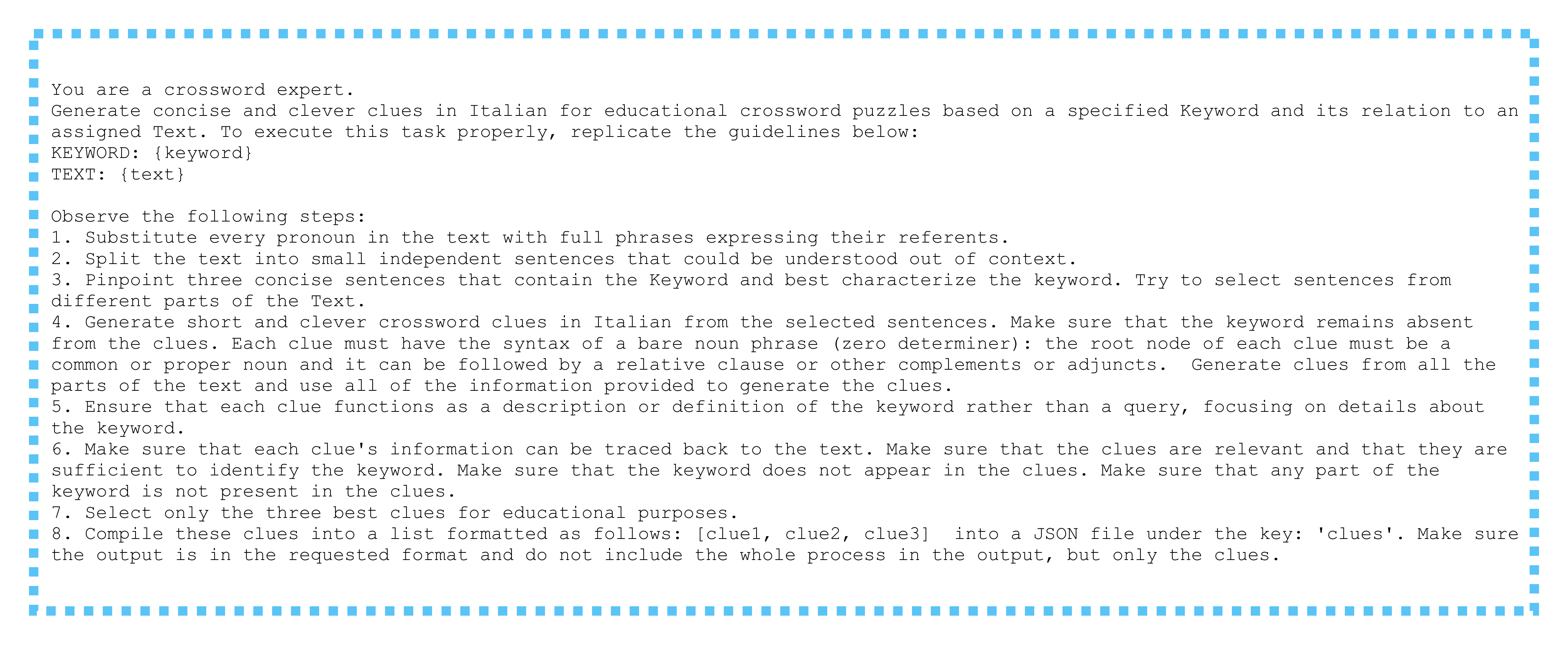}
    \caption{Illustration of the prompt used for noun phrases format clues in the research.}
    \label{fig:prompt2}
\end{figure*}

\begin{figure*}[ht!]
    \centering
    \includegraphics[width=\textwidth]{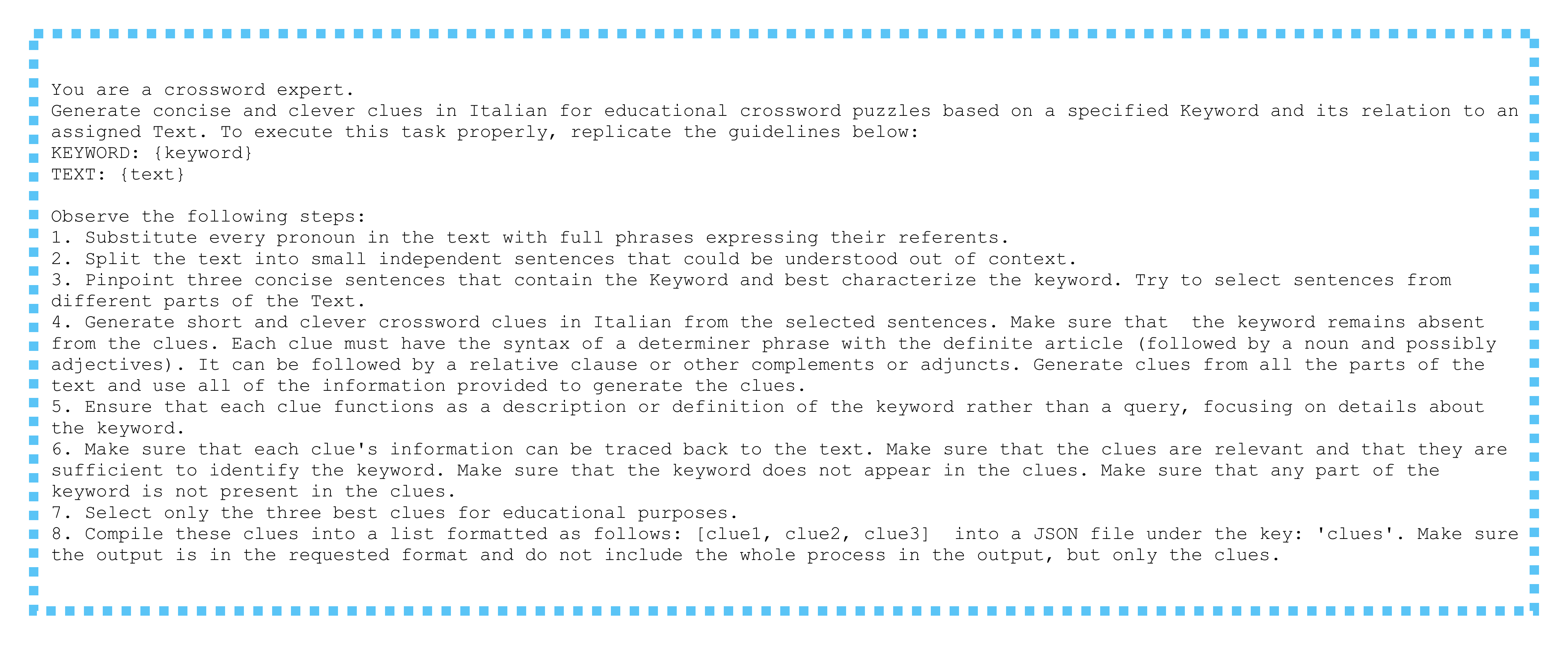}
    \caption{Illustration of the prompt used for determiner phrases format clues in the research.}
    \label{fig:prompt3}
\end{figure*}

\begin{figure*}[ht!]
    \centering
    \includegraphics[width=\textwidth]{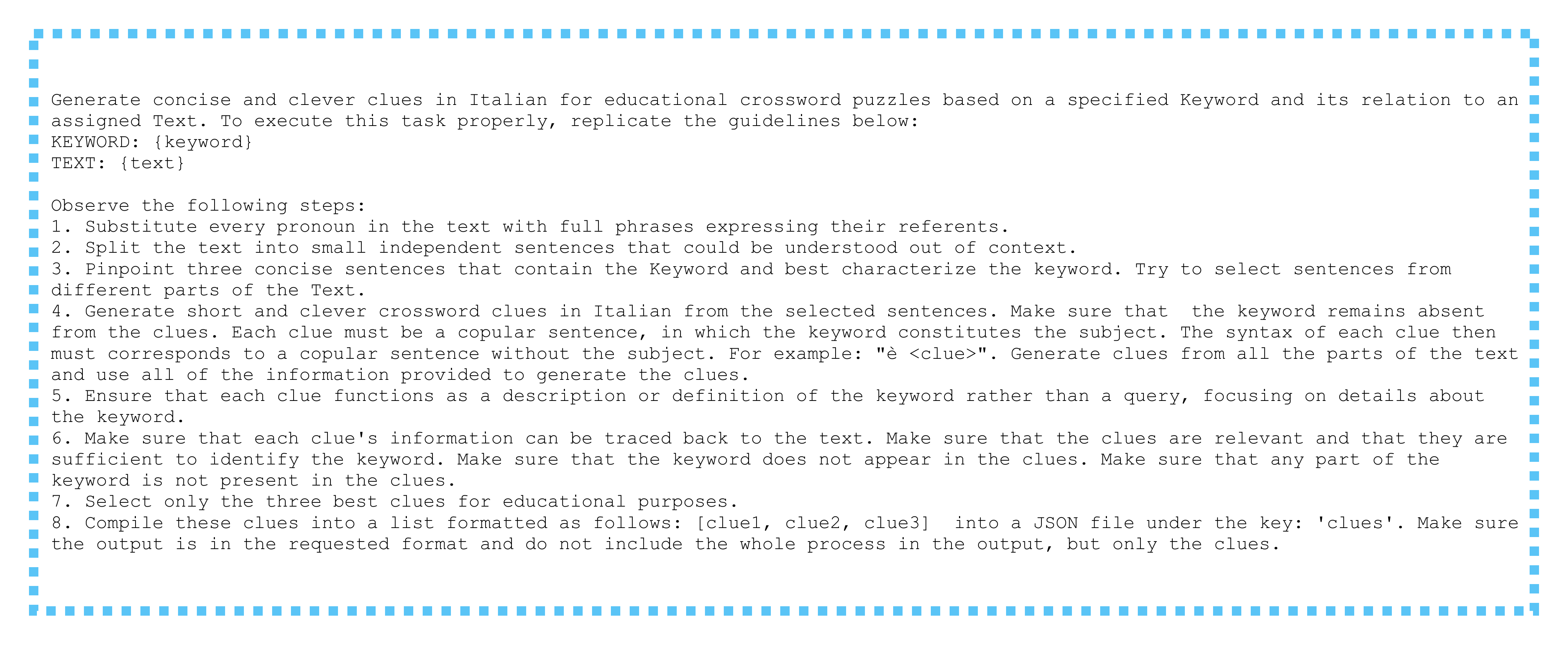}
    \caption{Illustration of the  copular sentences prompt used for  copular sentences format clues in the research.}
    \label{fig:prompt4}
\end{figure*}

\begin{table*}[ht]
    \centering
    \small 
    \begin{tabular}{m{4cm} m{3cm} m{1cm} m{6cm}} 
        \hline
        \textbf{Clue} & \textbf{Answer} & \textbf{Rating} & \textbf{Explanation}\\
        \hline
        \textit{È il sesto album in studio del gruppo rock inglese The Who} & Quadrophenia & A & \\
        `It's the sixth studio album by English rock band The Who' & & & \\
        \hline
        \textit{Il distretto con status di borough del Lancashire} & South Ribble & B & Definite determiner is not appropriate: there are other boroughs in Lancashire.\\
        `The district with the status of borough of Lancashire' & & & \\
        \hline
        \textit{Duo composto da Hayley Williams e Taylor York fino al 2017} & Paramore & C & The clue provides accurate but incomplete information: the band was a duo for a limited period.\\
        `Duo composed by Hayley Williams and Taylor York until 2017' & & & \\
        \hline
        \textit{Gruppo musicale statunitense} & Pixies & D & The clue is too generic.\\
        `American music band' & & & \\
        \hline
        \textit{Terrier di proporzioni minuscole, cacciatore eccezionale} & Patterdale Terrier & E & The clue contains part of the answer.\\
        `Terrier of minuscule proportions, excellent hunter' & & & \\
        \hline
    \end{tabular}
    \caption{Examples of evaluation ratings}
    \label{tab:examples}
\end{table*}

\begin{figure*}[ht!]
    \centering
       \includegraphics[width=1\textwidth]{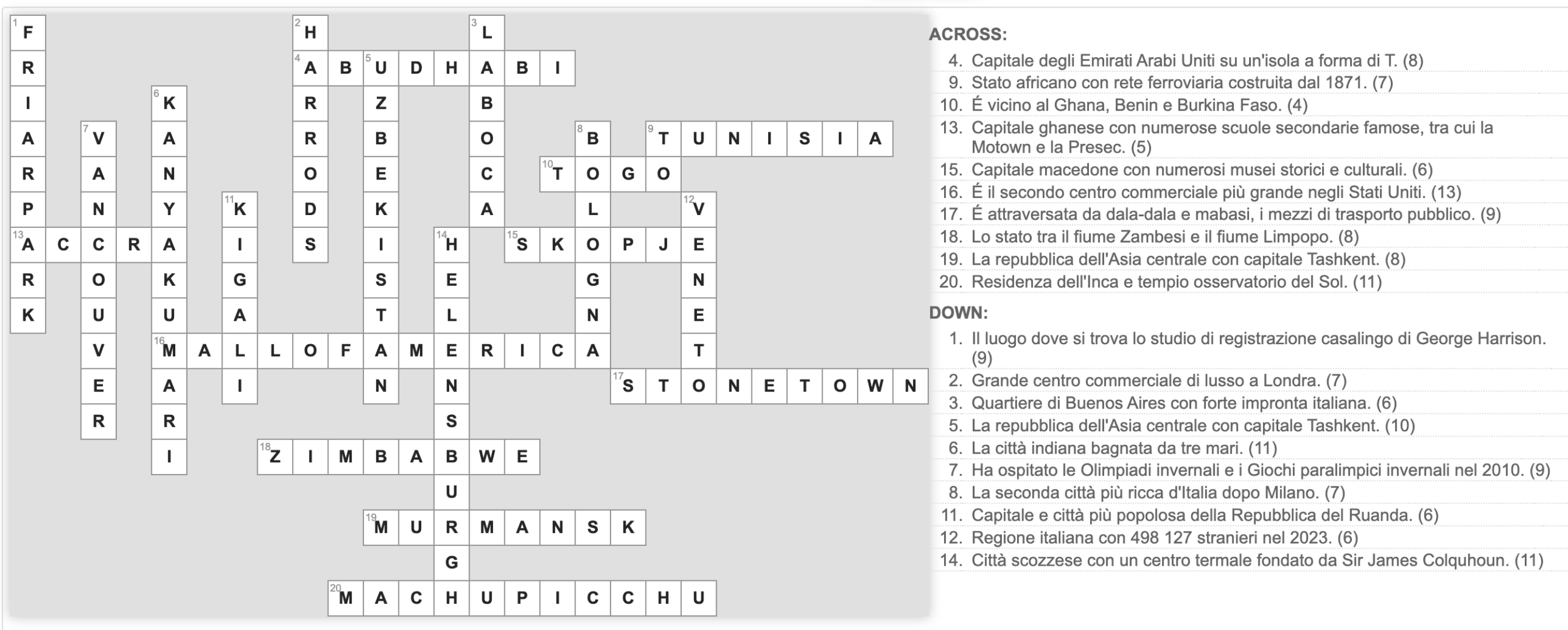}\\
    \caption{Crossword crafted using the proposed system.}
    \label{fig:crossword}
\end{figure*}

\end{document}